\documentclass[a4paper,conference]{IEEEtran}
\pdfoutput=1

\usepackage{amsmath,amssymb,amsfonts}
\usepackage{graphicx}
\usepackage{xcolor}
\usepackage{tikz}

\ifCLASSINFOpdf
\else
\fi
\usepackage{algorithmic}

\begin{document}
%
\title{JOADAA: joint online action detection and action anticipation}

\author{\IEEEauthorblockN{Mohammed Guermal, Rui Dai, Abid Ali, and François Brémond}
\IEEEauthorblockA{
Inria, Université Côte d'Azur, 2004 Route des Lucioles, 06902 Valbonne \\
\texttt{\{mohammed.guermal, rui.dai, francois.bremond\}@inria.fr}}}

%


\maketitle

\begin{abstract}
   Action anticipation involves forecasting future actions by connecting past events to future ones. However, this reasoning ignores the real-life hierarchy of events which is considered to be composed of three main parts: past, present, and future. We argue that considering these three main parts and their dependencies could improve performance.
On the other hand, online action detection is the task of predicting actions in a streaming manner. In this case, one has access only to the past and present information. Therefore, in online action detection (OAD) the existing approaches miss semantics or future information which limits their performance. To sum up, for both of these tasks, the complete set of knowledge (past-present-future) is missing, which makes it challenging to infer action dependencies, therefore having low performances. To address this limitation, we propose to fuse both tasks into a single uniform architecture. 
By combining action anticipation and online action detection, our approach can cover the missing dependencies of future information in online action detection. This method referred to as JOADAA, presents a uniform model that jointly performs action anticipation and online action detection. We validate our proposed model on three challenging datasets: THUMOS'14, which is a sparsely annotated dataset with one action per time step, CHARADES, and Multi-THUMOS, two densely annotated datasets with more complex scenarios. JOADAA achieves SOTA results on these benchmarks for both tasks.
\end{abstract}

\section{Introduction}
\label{sec:intro}
Envisioning upcoming occurrences plays a vital role in human intelligence as it aids in making choices while engaging with the surroundings. Humans possess an inherent skill to predict future happenings in diverse situations involving interactions with the environment. Likewise, the capacity to anticipate events is imperative for advanced AI systems operating in intricate settings, including interactions with other agents or individuals. The goal of online action detection (OAD) is to accurately pinpoint ongoing actions in streaming media, by predicting impending events. While action anticipation advances OAD and imitates the capacity of human cognition to anticipate events before they occur. Therefore, OAD and action anticipation are two important areas of research in computer vision, which have numerous applications in security surveillance, home-care, sports analysis, self-driving cars, and online danger detection. 
   Human perception of actions can be viewed as a continuous cycle in which prior knowledge is used to forecast future behavior, and then present knowledge is used to revise and update future predictions.
   To tackle action detection, we propose a unified framework of action anticipation and online action detection. Our predictions are in two steps, first we anticipate up-coming actions based on past information. Second, we update the anticipation by introducing the present information. By doing so, we gain in the online action detection by introducing the anticipated actions as pseudo-future information. In addition, it improves the action anticipation by comparing the prediction to the present information, thus combining them to improve both tasks.
    
    \begin{figure*}[!t]
\begin{center}
   \includegraphics[width=1\linewidth,height=6cm]{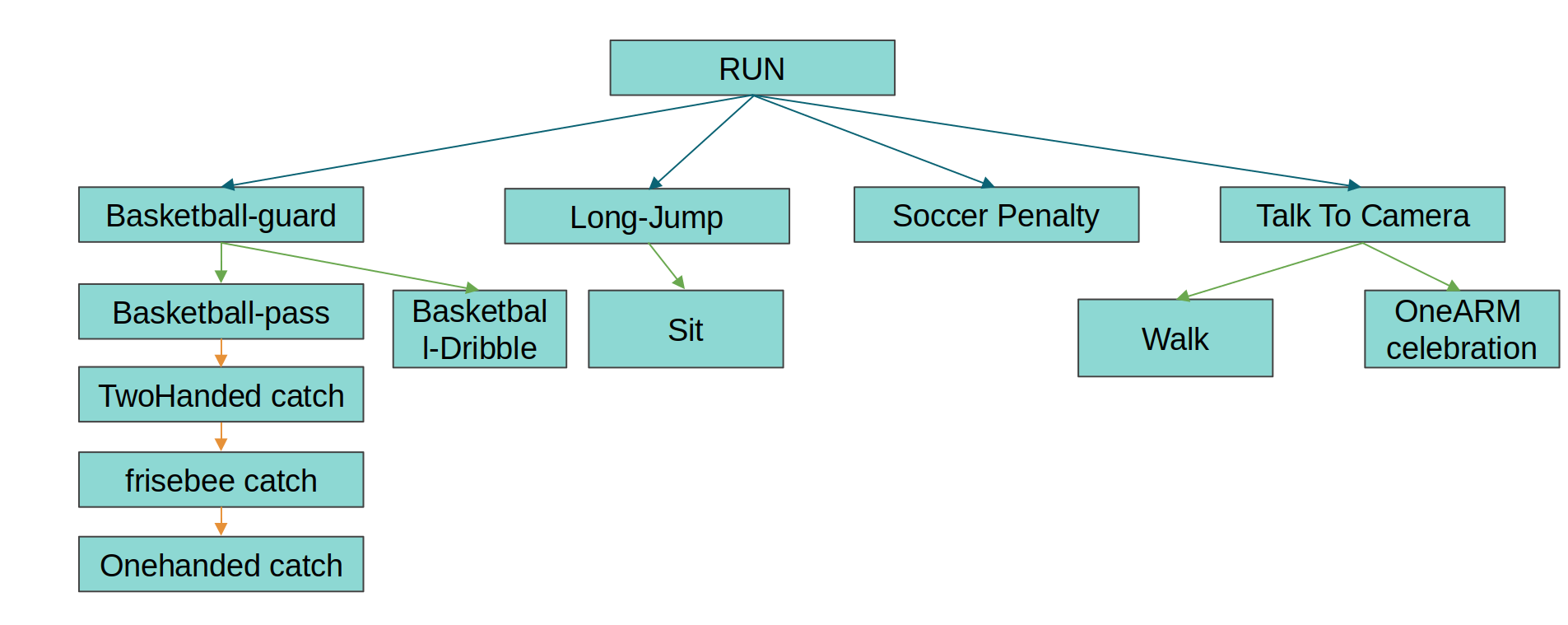}
   \caption{An example of human non-sequential dependencies. For instance, the actions \textit{RUN and OneHanded Catch} are highly correlated but distant. Also the same start action \textit{\textbf{RUN}} can lead to many different actions and scenarios. Therefore, it is very hard for online action detection or action anticipation to detect such relations without access to the future. In JOADAA, we propose to tackle this limitation by introducing a pseudo-future information by combining action anticipation and online action detection in the same task.} 
   \label{fig:data}
\end{center}
\end{figure*}
    
    Transformer networks such as \cite{attention,swin,vivit} have had a significant impact on computer vision and video understanding. This is due to their ability to capture long-range dependencies. LSTR \cite{LSTR}, TesTra \cite{TesTra}, or FUTR \cite{FUTR} have benefited from the transformer backbones to address the tasks of OAD and AA.   However, OAD and AA (action anticipation) tasks suffer from limited information as they don't have access to future information and global knowledge of the scene. This limited information restricts the ability of transformers to capture long-range dependencies and to learn significant relations between events. This can be demonstrated by comparing the effectiveness of models for offline action detection with online action detection. Offline, one has access to all pieces of information and a clear knowledge of the past, present, and future. Furthermore, complex densely annotated datasets (such as Multi-THUMOS \cite{multithumos}) have not been explored for online action detection and anticipation. It is challenging to recognize and foresee activities in such datasets. Most OAD architectures are only validated on sparsely-annotated activity datasets. Such simple annotated datasets are less challenging. First, these datasets do not have co-occurring actions. Second, they rarely have dependencies between actions in distant time steps. Furthermore, actions in densely annotated datasets have many possible outcomes. An example of these complex dependencies is given in Figure \ref{fig:data}. Due to these challenges, OAD methods are only validated on simple datasets. Therefore, even with the help of transformers, it is difficult to build knowledge of these long-range dependencies without having access to complete information.

    In the past, OAD and action anticipation have been treated as separate tasks. However, to tackle the above challenges, we propose JOADAA (Joint Online Action Detection and Action Anticipation) to tackle OAD and AA together. We create a pseudo-future when performing online action detection. By leveraging cross-attention between the real frame features and the anticipated frames, we enhance the quality of the features, thus improving the accuracy of the predictions by making the present aware of a pseudo-future. 
    Next, we propose to extract two types of information from these updated features: Local dependencies using TCNs (temporal convolution networks) and global dependencies using MHA (multi-head attention). Finally, we fuse both pieces of information to make online action detection predictions. 
    
    
    In this paper, following previous work, we extract features from video clips using 3D convolution neural networks (3D CNNs). We use I3D \cite{i3d} as a pre-trained backbone on the Kinetics dataset \cite{kinetics}. We store these extracted features in a memory bank. JOADAA consists of three main parts i) \textbf{Past Processing Block}, ii) \textbf{Anticipation prediction Block}, and iii) \textbf{Online action prediction Block}. First, we capture past information using a transformer encoder. The encoder output is first passed through a classification layer, which helps improve the quality of the embedding by making it class-dependent. Next, in the anticipation prediction part, we assume that we have not yet got the current frame. A transformer decoder is employed to learn from the last layer of the past embeddings to anticipate the upcoming actions in the next frame. This is carried out by introducing a set of learnable queries, called \textit{anticipation queries}. Finally, the online action prediction part uses anticipation embedding and current frame features to enhance the quality of the current frame. The new enhanced present frame features are fused with past features. Finally, global and local information is extracted using MHA and TCN layers, respectively, achieving a new enhanced feature map. Based on the challenges discussed, we propose the following main contributions:
    \begin{itemize}
        \item We propose a new architecture \textbf{JOADDA}, to jointly perform online action detection and action anticipation.
        \item We tackle both tasks for two different types of datasets, a densely annotated dataset and a simple activity dataset.
        \item We validate our proposed method on three benchmark datasets and achieve new SOTA results for online action detection and action anticipation. 
    \end{itemize}

\section{Related work}
\textbf{Online Action Detection} is the task of localizing action instances in time steps. We distinguish two types of action detection i.e., offline and online. In off-line action detection, the model has access to the entire video \cite{offline1,offline2,offline3,offline4,offline5}. Online action detection, on the other hand, occurs in real-time and has access to the past and the present only. RED \cite{red} uses reinforcement loss to encourage early recognition of activities. 
IDN \cite{idn} learns discriminative features and stores only knowledge that is relevant in the present. To achieve optimal features, LAP-Net \cite{lap} presents an adaptive sampling technique. PKD \cite{pkd} uses curriculum learning to transfer information from offline to online models. Shou et al. \cite{shou}, similar to early action detection, focus on online detection of action start (ODAS). StartNet \cite{startnet} divides ODAS into two stages and learns using a policy gradient. WOAD \cite{woad} employs video-level labeling and weakly-supervised learning. LSTR \cite{LSTR} uses a set of encoder-decoder architectures to capture the relations between long-term and short-term actions. They achieve state-of-the-art results on sparsely-annotated datasets but perform poorly on densely labeled datasets such as Multi-Thumas \cite{multithumos}.\\

\textbf{Action Anticipation} is the task of predicting future actions given the limited observation of a video. In the past, many strategies have been proposed to solve the next action anticipation, forecasting a single future action in a matter of seconds. Recently, the idea of anticipating long-term activities from a long-range video has been put out. Girdhar and Grauman \cite{girdhar} introduced the anticipative video transformer (AVT), which anticipates the following action using a self-attention decoder, which was further improved by FUTR \cite{FUTR} for minutes-long future actions. However, their architecture is suitable only for simple activities and simple datasets, which is not applicable to real-world scenarios that have multiple actions occurring at the same time.

Finally, in the study of mixing action anticipation and online action prediction, the authors in~\cite{TesTra} use the same architecture for both action anticipation and online action detection tasks. However, they dissociate these tasks, while we tackle both tasks jointly to improve both of them. Furthermore, the architecture in~\cite{TesTra} is very similar to \cite{LSTR}, therefore, the same limitations apply here as well.

In summary, to have adequate predictions, we need to build a well-descriptive hierarchy of information consisting of past, present, and future. Unfortunately, tasks such as online action detection or action anticipation do not have access to this global knowledge. In our work, we suggest combining OAD and AA in order to create pseudo-full knowledge that can improve action anticipation accuracy and produce comparable results for online action detection.
\begin{figure*}[!t]
\begin{center}
   \includegraphics[width=\linewidth]{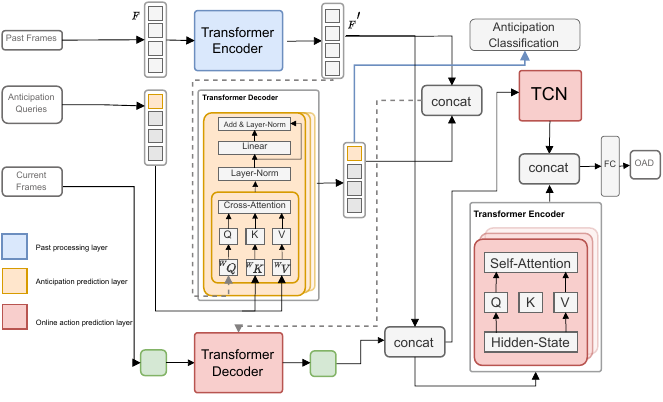}
   \end{center}
   \caption{Proposed JOADAA architecture with three units i) Past processing, ii) Anticipation prediction, and iii) Online Action prediction. Each stage is highlighted by a color for better understanding. Each block will  be explained in details in section \ref{method}} 
   \label{overview}
\end{figure*}
\section{Proposed method}\label{method}
The whole architecture consists of three main parts, i) Past Processing Block, ii) Anticipation prediction Block, and iii) Online Action Prediction, as shown in Figure \ref{overview}. First, a short-term past transformer-encoder enhances features. Second, an anticipation transformer-decoder anticipates the upcoming actions in the upcoming frames, using embedding output from the previous block and a set of learnable queries, which we call anticipation queries. Finally, a transformer-decoder uses the anticipation results and past information to predict the actions for the current frame (online action detection). Each module is explained in the following.

\subsection{Past Processing Block}
To enhance the ongoing action prediction, the initial stage in our model is to infer prior information. We employ a transformer encoder that accepts the embedding of previous frames as input. This enables us to highlight salient and robust frames by leveraging attention mechanisms, making our features more descriptive of previous activities (features). It can be challenging to identify which activity a person is performing solely based on the raw embedding or the current frame. For instance, if the current frame shows the person \textit{holding a bottle}, we are not sure if the ongoing action will be \textit{picking up the bottle, placing the bottle, drinking water, or pouring water}. However, if we know from the past that one of the previous actions was \textit{opening the bottle}, we can be more confident that the person is more likely to \textit{drink water}. These features are later used to anticipate future actions. Following \cite{attention}, the equations below sum up the first block of our architecture: 

\begin{equation} \label{eq:1}
F' = ATTENTION(F)
\end{equation}
\begin{equation} \label{eq:2}
ATTENTION(F) = Softmax(QK^{T}/\sqrt{d_k})V
\end{equation}
\begin{equation} \label{eq:3} 
    Q = W_q\times X , K = W_k \times X , V = W_v \times X
\end{equation}
\begin{equation}
    X = F + PE(F)
\end{equation}

PE stands for positional encoding, and $F$ $\in$ $\mathbb{R}^{T \times D}$ are the extracted features using the pre-trained I3D model \cite{i3d}, and $W_q$, $W_k$ and $W_v$ are learnable weights.

Furthermore, we propose different approaches for the use of past information. Following \cite{LSTR} we use long-term and short-term past information. Experimentally, the use of long-term and short-term past information is highly dependent on the type of dataset. The first intuition is that more information is always good for a neural network as it provides a more detailed description of events in a video. Especially with the use of transformers, we can capture long-range dependencies to learn all the steps that lead to the current actions. However, in our study, we find that this is not always true. For instance, the very long-past knowledge may sometimes harm performances, especially for densely annotated datasets. In scenarios where many actions co-occur, it is challenging to learn significant long-term relations, and thus these long-term features may act as noise to the model. Further experimental details are provided in Section \ref{ablation study}. 
 \subsection{Anticipation prediction Block}
 Inspired by \cite{FUTR}, the module takes a feature map $F^{'}$ $\in$ $\mathbb{R}^{T \times D}$  and a set of anticipation queries (learnable) $LQ$ $\in$ $\mathbb{R}^{N_q \times D}$, as inputs. Here, $N_q$ represents the number of queries and $D$ is the embedding dimension, which is the same as the feature map. Action anticipation can be achieved in two different ways.
 The first way is to proceed directly with a transformer encoder and to learn to predict the future. An encoder sees only a glimpse of the past and learns to predict the future. On the contrary, another way is to utilize a transformer decoder. In this approach, the strength of using learnable queries with a transformer decoder is that each query learns a specific feature for a specific frame in the future. The positional encoding indicates to the transformer the order of these learnable queries and helps the model relate each query to a corresponding point in the future. Additionally, by having these learnable queries in our model, it learns to adapt to each clip, since the queries are based on the past information of each clip. Therefore, these learnable queries learn to be aware of the past. JOADAA uses these learnable queries as a link between past events and possible future ones.
\begin{equation}\label{eq:5}
    N_q = 1 + N_f
\end{equation}
Where in Eq. \ref{eq:5}, 1 is for the upcoming frame that represents the ongoing action (represented in red in the Figure \ref{overview}). Since we do not have access yet to this frame; thus, it is also anticipated. $N_f$ is the number of frames to anticipate in the future to which we have no access. Information from the past, present, and future are connected by these learnable queries to improve both tasks efficiently. Later, these anticipation queries act as a pseudo-future to do the prediction of the ongoing action, see Section \ref{Online action prediction}.


\subsection{Online action prediction Block}\label{Online action prediction}
At this stage, we feed the features of the current frame and the previously learned features of potential actions in the current time step and subsequent time steps into a decoder. Our model can classify the current frame more accurately because it has pseudo-future knowledge. Modeling information this way has two effects. The prediction of the current frame is initially optimized by employing anticipation queries, and since we can access the current frame, we can also enhance the learned query on the current frame, which benefits our anticipation module. In addition, our local-to-global layers improve the performance of JOADAA. Adding a TCN layer (1D temporal convolution) helps the model capture local information. Transformers have proven to be a good tool to capture global and long-range dependencies. However, as explained earlier, this huge amount of information is not always helpful and may act as noise. Therefore, by mixing transformers with TCNs, our model learns complementary information from an updated feature map that we pass through an FC (fully connected) layer for classification. Notably, we utilize a Softmax layer for basic datasets with only one action at a time for validation and a Sigmoid layer for datasets with co-occurring actions in all categorization layers (past, future, and present).\\
Note that we use three different concatenation layers in our architecture. The first concatenation is between past frames features and anticipated frames features, the aim of this concatenation is to provide the decoder with a pseudo full information (past and pseudo future), which is the main idea of our paper (use AA to enhance OAD). The second concatenation is between past frames and the currently updated feature (since it is now aware of past and possible future actions). Here we only concatenate past and present because online action action detection is our main objective, which is why there is no more need for future information. The last concatenation is to use both local information learned through the TCNs and global information from the transformer decoder, which allows us to have better predictions as shown in the ablation studies Table \ref{tab:ab_LTG}.\\
We also use the same decoder for future frame anticipation and current frame prediction. Experiments have been conducted that showed that using different decoders does not improve the 
accuracy and sometimes leads to a slight decrease in accuracy. Hence, to keep the model lighter and have better prediction we keep the same weights. As for the encoders, the two of them are different; the last encoder is part of our proposed classification head, where we use a TCN to
capture local dependencies and a transformer encoder to capture long-range dependencies. Therefore, our intuition was not to share the weights between the encoders as they have a separate function in our architecture.
\section{Experiments}
In this section, we discuss experiments carried out for online action detection and action anticipation tasks on two different types of datasets. First, we briefly describe the datasets used and explain the implementation of the experiments carried out. Second, we compare JOADAA with existing SOTA methods for both online action detection and action anticipation. Finally, we explore the effectiveness of each module of our approach by performing an ablation study. More qualitative results are provided in the supplementary materials.
\subsection{Datasets}
In this section, we briefly explain the datasets used in our experiments. We experiment on two types of datasets, i) sparsely annotated dataset (THUMOS'14 \cite{thumos}), and ii) densely annotated datasets (Multi-THUMOS \cite{multithumos} and CHARADES \cite{charades}). Each of them is described below.

.\textbf{THUMOS'14}: contains 413 untrimmed videos with 20 categories of actions. The dataset is divided into two subsets: the validation set and the test set. The validation set contains 200 videos, and the test set contains 213 videos. Following common practice, we use the validation set for training and report the results in the test set. More details are available in \cite{thumos}.

\textbf{Multi-THUMOS}: contains dense, multilabel frame-level action annotations for 30 hours across 400 videos from the THUMOS'14 \cite{thumos} action detection dataset. It consists of 38,690 annotations of 65 action classes, with an average of 1.5 labels per frame and 10.5 action classes per video. More details can be found in \cite{multithumos}.

\textbf{CHARADES}: is composed of 9,848 videos of daily indoor activities with an average length of 30 seconds, involving interactions with 46 object classes in 15 types of indoor scenes and containing a vocabulary of 30 verbs leading to 157 action classes. Readers can find more details in \cite{charades}.\\

\subsection{Implementation Details}
We implement our proposed model in PyTorch \cite{pytorch}. All experiments are performed on a system with 3 Nvidia V100 graphics cards. For all Transformer units, we set their number of heads to 16 and hidden units to 1024 dimensions. To learn the weights of the model, we use Adam Optimizer \cite{adam} with weight decay $5 \times 10^{-5}$. The learning rate increases linearly from zero to $5 \times 10^{-5}$ in the first 40\% training iterations and then decreases to zero using a cosine warm-up. Our models are optimized with a batch size of 16, and trained for 25 epochs.
\noindent\textbf{Evaluation protocol}: We follow previous work and use mean average precision per frame (mAP) to evaluate performances.
\begin{table*}[h]
\centering
\begin{tabular}{cccc}
\hline
\multicolumn{1}{l}{}                 & THUMOS'14                         & \multicolumn{1}{l}{Multi-THUMOS} & CHARADES                       \\ \hline
FATS\cite{fats}     & 59.0                           & -                                & -                              \\
IDN\cite{idn}       & 60.3                           & -                                & -                              \\
PKD\cite{pkd}       & 64.5                           & -                                & -                              \\
WOAD\cite{woad}     & 67.1                           & -                                & -                              \\
LFB\cite{lfb}       & 64.8                           & -                                & -                              \\
TRN\cite{trn}       & 62.1                           & 39.5                             & 18.3                              \\
PDAN\cite{offline5}                               & 62.2                           & 32.6                             & 16.0                           \\
MSTCT\cite{mstct}                               & 70.5                           & 41.4                             & 19.5                           \\
LSTR\cite{LSTR}     & 69.5                           & 43.0                             & 20.0                           \\
TesTra\cite{TesTra} & 71.2                           & 41.7                             & 19.9                           \\
GateHUB\cite{gatehub}                              & 70.7                           & -                                & -                              \\
\multicolumn{1}{l}{JOADAA}           & \textbf{72.6} & \textbf{45.2}   & \textbf{21.5} \\ \hline
\end{tabular}
\caption{State of the art comparison for OAD on THUMOS'14, Multi-THUMOS, and CHARADES. Due to the lack of available OAD methods for CHARADES and Multi-THUMOS datasets, we compare also with two off-line methods PDAN and MSTCT, adapted to an online setting.}
\label{tab:state of the art results comparison for OAD task}
\end{table*}

\begin{table*}[h!]

\begin{center}

\begin{tabular}{cclllcllcll}

\hline
\multicolumn{1}{l}{} & \multicolumn{4}{c}{THUMOS'14}                                                                 & \multicolumn{3}{l}{Multi-THUMOS}                                   & \multicolumn{3}{c}{CHARADES}                          \\ \hline
\multicolumn{1}{l}{} & \multicolumn{1}{l}{1} & 2             & 4             & \multicolumn{1}{l|}{6}             & 2             & 4             & \multicolumn{1}{l|}{6}             & \multicolumn{1}{l}{2} & 4             & 6             \\
TTM\cite{ttm}                 & 46.8                  & 45.5          & 43.6          & \multicolumn{1}{l|}{41.1}          & -             & -             & \multicolumn{1}{l|}{-}             & -                     & -             & -             \\
LSTR\cite{LSTR}                 & 60.4                  & 58.6          & 53.3          & \multicolumn{1}{l|}{48.9}          & -             & -             & \multicolumn{1}{l|}{-}             & -                     & -             & -             \\
TesTra\cite{TesTra}               & 66.2                  & 63.5          & 57.4          & \multicolumn{1}{l|}{52.6}          & 28.0          & 22.4          & \multicolumn{1}{l|}{19.8}          & 18.1                  & 13.7          & 13.5          \\
JOADAA               & \textbf{67.7}         & \textbf{63.9} & \textbf{62.9} & \multicolumn{1}{l|}{\textbf{59.3}} & \textbf{42.5} & \textbf{37.7} & \multicolumn{1}{l|}{\textbf{35.2}} & \textbf{20.2}         & \textbf{19.5} & \textbf{19.0} \\ \hline
\end{tabular}

\end{center}
\caption{Comparison with SOTA for the action anticipation task. 1, 2, 4, and 6 represent the number of anticipated frames. We notice that our method is more robust w.r.t. the number of anticipated frames compared to other methods where accuracy drops dramatically.}
\label{tab:sota anticipation}
\end{table*}

\begin{table}[h]
\begin{center}
\resizebox{\columnwidth}{!}{
\begin{tabular}{lccll}
\hline
   Dataset               & 1                                  & 2                              & 4         & 6         \\ \hline
THUMOS'14            & { \textcolor{blue}{70.5} / \textcolor{red}{67.7}} & \textcolor{blue}{71.5} / \textcolor{red}{63.9}                      & \textcolor{blue}{72.2} / \textcolor{red}{62.9} & \textcolor{blue}{72.6} / \textcolor{red}{59.3} \\ 
CHARADES & \textcolor{blue}{20.0} / \textcolor{red}{20.7}                & \textcolor{blue}{21.4} / \textcolor{red}{20.2}             & \textcolor{blue}{21.5} / \textcolor{red}{19.5} & \textcolor{blue}{21.4} / \textcolor{red}{19.0} \\ 
Multi-THUMOS      & \multicolumn{1}{l}{\textcolor{blue}{44.5} / \textcolor{red}{42.8}}     & \multicolumn{1}{l}{\textcolor{blue}{45.2} / \textcolor{red}{42.5}} & \textcolor{blue}{45.0} / \textcolor{red}{37.7} & \textcolor{blue}{45.2} / \textcolor{red}{35.2} \\ \hline
\end{tabular}
}
\end{center}
\caption{Effect of \textcolor{red}{action anticipation} prediction and \textcolor{blue}{online action detection} using long-short-term knowledge. 1, 2, 4, and 6 are the number of anticipated frames. Best viewed in color.}
\label{tab:anticip further}
\end{table}

\begin{table}[h] 
\begin{center}

\begin{tabular}{lcll}
\hline
      Dataset            & 2                              & 4         & 6         \\ \hline
THUMOS'14            & \textcolor{blue}{70.6} / \textcolor{red}{64.4}                      & \textcolor{blue}{70.0} / \textcolor{red}{63.0} & \textcolor{blue}{70.6} / \textcolor{red}{58.2} \\ 
CHARADES & \textcolor{blue}{21.8} / \textcolor{red}{20.4}            & \textcolor{blue}{21.4} / \textcolor{red}{19.5} & \textcolor{blue}{21.3} / \textcolor{red}{19.0} \\ 
Multi-THUMOS      & \multicolumn{1}{l}{\textcolor{blue}{45.1} / \textcolor{red}{36.9}} & \textcolor{blue}{45.3} / \textcolor{red}{39.2} & \textcolor{blue}{45.1} / \textcolor{red}{37.3} \\ \hline
\end{tabular}
\end{center}
\caption{Results of using only short-term past information on multiple datasets for \textcolor{blue}{online action detection} and \textcolor{red}{action anticipation}. 2, 4, and 6 are the number of anticipated frames.}
\label{tab:short_term}
\end{table}

\begin{table}[h]
\begin{center}
\resizebox{\columnwidth}{!}{
\begin{tabular}{lcc|cc}
\hline
Dataset      & \multicolumn{2}{c|}{long term past + short term past} & \multicolumn{2}{c}{short term past} \\ \cline{2-5} 
             & LSTR                     & JOADAA                     & LSTR            & JOADAA            \\ \hline
THUMOS'14       & 69.5                     & \textbf{72.6}                       & 65.4            & \textbf{70.6}              \\
Multi-THUMOS & 42.0                     & \textbf{45.2}                       & 40.0            & \textbf{45.1}              \\
CHARADES     & 20.0                     & \textbf{21.4}                       & 19.8            & \textbf{21.3}              \\ \hline
\end{tabular}
}
\end{center}
\caption{Comparison of JOADAA with LSTR method using long-past information. JOADAA is more robust to utilize long-past information.}
\label{tab:ab short lstr}
\end{table}



\subsection{Comparison with the SoTA}
\subsubsection{OAD Comparison on the simple dataset (THUMOS'14)}
Table \ref{tab:state of the art results comparison for OAD task} presents the results of online action detection. For the THUMOS'14 \cite{thumos} dataset we achieve state-of-the-art results by a margin of \textbf{1.4\%}. GateHUB\cite{gatehub} was SoTA results for OAD on the THUMOS'14 dataset. However, they provide two results on this dataset, one with TSN as the backbone feature extractor and one with Timesformer\cite{tf}. Upon careful examination, we noticed the following points: 1) Our accuracy still surpasses theirs. 2) The GateHUB method was not compared with TesTra, which demonstrated better accuracy with the same settings. 3) GateHUB achieves SOTA results only when TimeSformer\cite{tf} is used as an RGB feature extractor, making it difficult to determine whether the results are due to the extractor or to their proposed solution. In conclusion, while the GateHUB paper argues for capturing relevant information from the past to the present, our JOADAA method, which employs a simple implementation of transformers, outperforms it along with TesTra\cite{TesTra}.
\subsubsection{OAD comparison on densely annotated datasets}
We evaluate JOADAA on more complex datasets such as Multi-THUMOS\cite{multithumos} and CHARADES \cite{charades}. We utilize LSTR \cite{LSTR}, TesTra\cite{TesTra}, and TRN\cite{trn} to train on these datasets to build baseline methods, as there are no validated online methods to compare JOADAA to these datasets. JOADAA improves the baselines by \textbf{1.5\%} on CHARADES\cite{charades} and \textbf{2.2\%} on Multi-THUMOS \cite{multithumos} dataset. The main difference between our approach and baseline methods \cite{LSTR} and \cite{TesTra}, is the introduction of pseudo-future knowledge to our online action prediction. It helps make more precise predictions by having a knowledge of different possible outcomes.
\subsubsection{OAD comparison using off-line methods}
For further comparison, we adapt offline methods to online settings. We use PDAN\cite{offline5} and MSTCT\cite{mstct} two SoTA methods on CHARADES and Multi-THUMOS in off-line action detection. We outperform these two methods on all three datasets THUMOS'14, Multi-THUMOS, and CHARADES.
\subsubsection{AA SoTA comparison}
Similarly, our model achieves SOTA results on action anticipation as noted in Table \ref{tab:sota anticipation}. When Increasing the anticipated frames from 1 to 6, TesTra's \cite{TesTra} accuracy drops by \textbf{13.6\%} on the THUMOS'14 dataset, whereas our model decreases by only \textbf{8.4\%}, which showcases robustness of our proposed solution. Also, JOADAA performs much better in more complex datasets (CHARADES and Multi-THUMOS).

In Table \ref{tab:anticip further}, we demonstrate how far we can foresee the future. We notice that, in general, the further we anticipate, the better the accuracy of the online action detection (\textcolor{blue}{blue}) until it reaches a level where the accuracy stops increasing. Such a behavior makes sense because the model can learn more action dependencies by inferring more information about upcoming events. On the other hand, action anticipation results (\textcolor{red}{red}) decrease when the anticipation period increases, because the model has more space to explore.

\subsection{Ablation study}\label{ablation study}
In this section, we discuss how the different modules contribute to JOADAA. 
\subsubsection{Ablation on the past processing block}
\begin{table}[h]
\centering
\begin{tabular}{c|c}
\hline
    Module                & THUMOS'14 \\ \hline
Transformer encoder & 71.5      \\ \hline
LSTM+Conv           & 54.2      \\ \hline
\end{tabular}
\caption{Comparing two techniques for past information processing. We use a transformer encoder and a set of LSTM blocks with a convolution layer.}
\label{tab:past_lstm}
\end{table}
First, we analyze the use of long-range past features on different datasets. As discussed in Section \ref{method}, past information can be used in two manners, either using only short-term past (32 frames) or long-short-term past (512+32 frames). This past information is used to infer the pseudo-future in our approach. In Tables \ref{tab:short_term} and \ref{tab:ab short lstr}, we observe that our model is more robust when it comes to using only short-term past information (decreases by \textbf{2\%}) on the THUMOS'14\cite{thumos}, unlike LSTR \cite{LSTR} where the accuracy decreases by \textbf{4.1\%}. 
One important result of our study is that long-past knowledge is more important for simple actions (single-action datasets) than for complex actions (densely annotated datasets). This is because numerous actions may occur simultaneously without being connected in densely annotated datasets, making it more challenging to infer relations from them. As a result, including information from the distant past can skew model predictions.\\

Recently, transformers have been widely used, since they outperformed the existing approaches such as 3D-CNNs and RNNs. In fact, 3D-CNNs are known to be good general feature extractors as they can capture overall visual appearances in a video. However, their CNN filters capture pixel-level information in a local neighborhood but struggle with long-term dependencies. Therefore, we limit the use of 3D-CNNs to extract video clip features for our architecture. Furthermore, action detection tasks require a strong grasp of long-range temporal dependencies, and transformers excel at capturing long-term information compared to RNNs. Therefore, the transformers are the best choice for OAD and AA tasks. However, most papers lately use transformers based on the previous intuition without any justification.\\

Table \ref{tab:past_lstm}  presents a comparison study between RNNs (LSTMs\cite{lstm}) and transformers. We replace our first encoder for past information processing with 3 blocks of LSTM and a convolution layer to reduce the feature map size. Results show that transformers are better suited for capturing long-range dependencies and produce far more better results which justifies our design choice.
\subsubsection{Ablation on the action anticipation module}
\begin{table}[h]
\begin{center}

\begin{tabular}{lcc}
\hline
Dataset & OAD+AA & OAD  \\ \hline
THUMOS'14  & 72.6   & 71.2 \\ \hline
\end{tabular}

\end{center}
\caption{Analyzing the JOADAA behavior with and without action anticipation.}
\label{tab:ab_AA}
\end{table}
Another ablation study is done in Table \ref{tab:ab_AA}. We conduct two main experiments: one with the full JOADAA model and the other one without the Action Anticipation (AA) module. We can see that the AA module enhances online action detection, which supports our claim that combining AA and OAD leads to better results.

\subsubsection{Ablation on the OAD prediction layer}
\begin{table}[h]
\begin{center}
\begin{tabular}{lcc}
\hline
Dataset & TCN+TR. Encoder & FC  \\ \hline
THUMOS'14  & 72.6           & 69.7 \\ \hline
\end{tabular}
\end{center}
\caption{Effect of fusing local and global information on OAD. FC stands for fully-connected layer. As expected capturing different type of dependencies provides better results.}
\label{tab:ab_LTG}
\end{table}
Table \ref{tab:ab_LTG} shows the effect of fusing local and global knowledge, in contrast to using directly the output of the decoder on the current frame which carries only global information in it. By doing so, our results increase by \textbf{2.9\%}. As argued earlier, this is due to the fact that TCNs can extract local changes and better detect relations in neighboring frames, whereas baseline transformers capture long-range dependencies that sometimes are not adapted to predicting the current frame events.

\subsection{Qualitative Analysis}
\begin{figure}[h]
 \begin{center}
    \includegraphics[width=\linewidth,height=5cm]{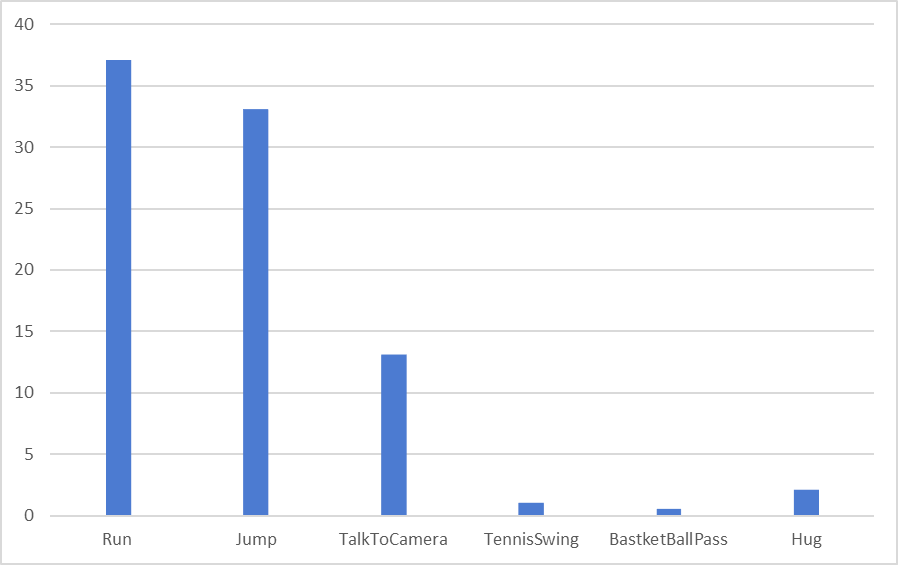}
    \caption{Action anticipation accuracy improvement on six actions w.r.t. TesTra  model. This is performed on the Multi-THUMOS dataset, using 4 frames as anticipation length.}
    \label{fig:qualitative}
\end{center}
\end{figure}
In this section, we analyze the effectiveness of our method on densely annotated datasets. We study anticipation improvement on six different actions, from the Multi-THUMOS dataset, according to their complexity as shown in Figure \ref{fig:qualitative}. We observe that the gain in some of these actions can reach 37\%, while in some other actions, it is almost zero.

In fact, our prediction block anticipates the upcoming frame alongside future frames. By having access to the current frame our model can correlate the anticipated action to the real action, hence we can learn to better anticipate the current frame, leading to a better-performing anticipation module.

Upon closer examination of these actions, we find that the improvement is particularly important for activities where there are multiple dependencies, or if the activity is interconnected with many other actions. The action \textbf{Run} for instance, has correlations with up to seven other activities, as illustrated in Figure \ref{fig:data}. 

The qualitative results in Figure \ref{fig:qualitative} demonstrate the robustness of JOADAA for complex correlated activities. This opens doors for future studies to analyze OAD and action anticipation on complex dense datasets.

\section{Conclusion} 
Online action detection and anticipation are important fields in computer vision that have many real-world applications. These two tasks are highly correlated, and that is why we design JOADAA to address both tasks jointly, improving one using the other and vice versa. Furthermore, we discuss the limitations of OAD and action anticipation for sparsely and densely annotated datasets.\\

Our model is limited in terms of effectively using long-range past features, especially for densely annotated datasets. Past knowledge undoubtedly adds to current knowledge and should lead to improvements. However, as demonstrated in this study, just adding pre-extracted features to transformers can also introduce noise. In the future, we are interested in tackling this limitation by modeling past features more effectively. One possible solution is to use an intermediate filter to learn only important features \cite{ctrn} or to learn the dependencies using a graph model to model only relevant features following \cite{thorn}.

\newpage
\bibliographystyle{IEEEtran}
\bibliography{IEEEabrv}

\begin{thebibliography}{10}
\providecommand{\url}[1]{#1}
\csname url@samestyle\endcsname
\providecommand{\newblock}{\relax}
\providecommand{\bibinfo}[2]{#2}
\providecommand{\BIBentrySTDinterwordspacing}{\spaceskip=0pt\relax}
\providecommand{\BIBentryALTinterwordstretchfactor}{4}
\providecommand{\BIBentryALTinterwordspacing}{\spaceskip=\fontdimen2\font plus
\BIBentryALTinterwordstretchfactor\fontdimen3\font minus \fontdimen4\font\relax}
\providecommand{\BIBforeignlanguage}[2]{{%
\expandafter\ifx\csname l@#1\endcsname\relax
\typeout{** WARNING: IEEEtran.bst: No hyphenation pattern has been}%
\typeout{** loaded for the language `#1'. Using the pattern for}%
\typeout{** the default language instead.}%
\else
\language=\csname l@#1\endcsname
\fi
#2}}
\providecommand{\BIBdecl}{\relax}
\BIBdecl

\bibitem{attention}
A.~Vaswani, N.~Shazeer, N.~Parmar, J.~Uszkoreit, L.~Jones, A.~N. Gomez, {\L}.~Kaiser, and I.~Polosukhin, ``Attention is all you need,'' \emph{Advances in neural information processing systems}, vol.~30, 2017.

\bibitem{swin}
Z.~Liu, J.~Ning, Y.~Cao, Y.~Wei, Z.~Zhang, S.~Lin, and H.~Hu, ``Video swin transformer,'' in \emph{Proceedings of the IEEE/CVF Conference on Computer Vision and Pattern Recognition}, 2022, pp. 3202--3211.

\bibitem{vivit}
A.~Arnab, M.~Dehghani, G.~Heigold, C.~Sun, M.~Lu{\v{c}}i{\'c}, and C.~Schmid, ``Vivit: A video vision transformer,'' in \emph{Proceedings of the IEEE/CVF International Conference on Computer Vision}, 2021, pp. 6836--6846.

\bibitem{LSTR}
M.~Xu, Y.~Xiong, H.~Chen, X.~Li, W.~Xia, Z.~Tu, and S.~Soatto, ``Long short-term transformer for online action detection,'' \emph{Advances in Neural Information Processing Systems}, vol.~34, pp. 1086--1099, 2021.

\bibitem{TesTra}
Y.~Zhao and P.~Kr{\"a}henb{\"u}hl, ``Real-time online video detection with temporal smoothing transformers,'' in \emph{European Conference on Computer Vision}.\hskip 1em plus 0.5em minus 0.4em\relax Springer, 2022, pp. 485--502.

\bibitem{FUTR}
D.~Gong, J.~Lee, M.~Kim, S.~J. Ha, and M.~Cho, ``Future transformer for long-term action anticipation,'' in \emph{Proceedings of the IEEE/CVF Conference on Computer Vision and Pattern Recognition}, 2022, pp. 3052--3061.

\bibitem{multithumos}
S.~Yeung, O.~Russakovsky, N.~Jin, M.~Andriluka, G.~Mori, and L.~Fei-Fei, ``Every moment counts: Dense detailed labeling of actions in complex videos,'' \emph{International Journal of Computer Vision}, vol. 126, no.~2, pp. 375--389, 2018.

\bibitem{i3d}
J.~Carreira and A.~Zisserman, ``Quo vadis, action recognition? a new model and the kinetics dataset,'' in \emph{proceedings of the IEEE Conference on Computer Vision and Pattern Recognition}, 2017, pp. 6299--6308.

\bibitem{kinetics}
W.~Kay, J.~Carreira, K.~Simonyan, B.~Zhang, C.~Hillier, S.~Vijayanarasimhan, F.~Viola, T.~Green, T.~Back, P.~Natsev \emph{et~al.}, ``The kinetics human action video dataset,'' \emph{arXiv preprint arXiv:1705.06950}, 2017.

\bibitem{offline1}
Z.~Shou, D.~Wang, and S.-F. Chang, ``Temporal action localization in untrimmed videos via multi-stage cnns,'' in \emph{Proceedings of the IEEE conference on computer vision and pattern recognition}, 2016, pp. 1049--1058.

\bibitem{offline2}
Z.~Shou, J.~Chan, A.~Zareian, K.~Miyazawa, and S.-F. Chang, ``Cdc: Convolutional-de-convolutional networks for precise temporal action localization in untrimmed videos,'' in \emph{Proceedings of the IEEE conference on computer vision and pattern recognition}, 2017, pp. 5734--5743.

\bibitem{offline3}
H.~Xu, A.~Das, and K.~Saenko, ``R-c3d: Region convolutional 3d network for temporal activity detection,'' in \emph{Proceedings of the IEEE international conference on computer vision}, 2017, pp. 5783--5792.

\bibitem{offline4}
Y.~Zhao, Y.~Xiong, L.~Wang, Z.~Wu, X.~Tang, and D.~Lin, ``Temporal action detection with structured segment networks,'' in \emph{Proceedings of the IEEE International Conference on Computer Vision}, 2017, pp. 2914--2923.

\bibitem{offline5}
R.~Dai, S.~Das, L.~Minciullo, L.~Garattoni, G.~Francesca, and F.~Bremond, ``Pdan: Pyramid dilated attention network for action detection,'' in \emph{Proceedings of the IEEE/CVF Winter Conference on Applications of Computer Vision}, 2021, pp. 2970--2979.

\bibitem{red}
J.~Gao, Z.~Yang, and R.~Nevatia, ``Red: Reinforced encoder-decoder networks for action anticipation,'' \emph{arXiv preprint arXiv:1707.04818}, 2017.

\bibitem{idn}
H.~Eun, J.~Moon, J.~Park, C.~Jung, and C.~Kim, ``Learning to discriminate information for online action detection,'' in \emph{Proceedings of the IEEE/CVF conference on computer vision and pattern recognition}, 2020, pp. 809--818.

\bibitem{lap}
S.~Qu, G.~Chen, D.~Xu, J.~Dong, F.~Lu, and A.~Knoll, ``Lap-net: Adaptive features sampling via learning action progression for online action detection,'' \emph{arXiv preprint arXiv:2011.07915}, 2020.

\bibitem{pkd}
P.~Zhao, L.~Xie, Y.~Zhang, Y.~Wang, and Q.~Tian, ``Privileged knowledge distillation for online action detection,'' \emph{arXiv preprint arXiv:2011.09158}, 2020.

\bibitem{shou}
Z.~Shou, J.~Pan, J.~Chan, K.~Miyazawa, H.~Mansour, A.~Vetro, X.~G. Nieto, and S.-F. Chang, ``Online action detection in untrimmed, streaming videos-modeling and evaluation,'' in \emph{ECCV}, vol.~1, no.~2, 2018, p.~5.

\bibitem{startnet}
M.~Gao, M.~Xu, L.~S. Davis, R.~Socher, and C.~Xiong, ``Startnet: Online detection of action start in untrimmed videos,'' in \emph{Proceedings of the IEEE/CVF International Conference on Computer Vision}, 2019, pp. 5542--5551.

\bibitem{woad}
M.~Gao, Y.~Zhou, R.~Xu, R.~Socher, and C.~Xiong, ``Woad: Weakly supervised online action detection in untrimmed videos,'' in \emph{Proceedings of the IEEE/CVF Conference on Computer Vision and Pattern Recognition}, 2021, pp. 1915--1923.

\bibitem{girdhar}
R.~Girdhar and K.~Grauman, ``Anticipative video transformer,'' in \emph{Proceedings of the IEEE/CVF international conference on computer vision}, 2021, pp. 13\,505--13\,515.

\bibitem{thumos}
H.~Idrees, A.~R. Zamir, Y.-G. Jiang, A.~Gorban, I.~Laptev, R.~Sukthankar, and M.~Shah, ``The thumos challenge on action recognition for videos “in the wild”,'' \emph{Computer Vision and Image Understanding}, vol. 155, pp. 1--23, 2017.

\bibitem{charades}
J.~Zhang, F.~Shen, X.~Xu, and H.~T. Shen, ``Temporal reasoning graph for activity recognition,'' \emph{IEEE Transactions on Image Processing}, vol.~29, pp. 5491--5506, 2020.

\bibitem{pytorch}
A.~D.~I. Pytorch, ``Pytorch,'' 2018.

\bibitem{adam}
D.~P. Kingma and J.~Ba, ``Adam: A method for stochastic optimization,'' \emph{arXiv preprint arXiv:1412.6980}, 2014.

\bibitem{fats}
Y.~H. Kim, S.~Nam, and S.~J. Kim, ``Temporally smooth online action detection using cycle-consistent future anticipation,'' \emph{Pattern Recognition}, vol. 116, p. 107954, 2021.

\bibitem{lfb}
C.-Y. Wu, C.~Feichtenhofer, H.~Fan, K.~He, P.~Krahenbuhl, and R.~Girshick, ``Long-term feature banks for detailed video understanding,'' in \emph{Proceedings of the IEEE/CVF Conference on Computer Vision and Pattern Recognition}, 2019, pp. 284--293.

\bibitem{trn}
M.~Xu, M.~Gao, Y.-T. Chen, L.~S. Davis, and D.~J. Crandall, ``Temporal recurrent networks for online action detection,'' in \emph{Proceedings of the IEEE/CVF International Conference on Computer Vision}, 2019, pp. 5532--5541.

\bibitem{mstct}
R.~Dai, S.~Das, K.~Kahatapitiya, M.~S. Ryoo, and F.~Br{\'e}mond, ``Ms-tct: multi-scale temporal convtransformer for action detection,'' in \emph{Proceedings of the IEEE/CVF Conference on Computer Vision and Pattern Recognition}, 2022, pp. 20\,041--20\,051.

\bibitem{gatehub}
J.~Chen, G.~Mittal, Y.~Yu, Y.~Kong, and M.~Chen, ``Gatehub: Gated history unit with background suppression for online action detection,'' in \emph{Proceedings of the IEEE/CVF Conference on Computer Vision and Pattern Recognition}, 2022, pp. 19\,925--19\,934.

\bibitem{ttm}
X.~Wang, S.~Zhang, Z.~Qing, Y.~Shao, Z.~Zuo, C.~Gao, and N.~Sang, ``Oadtr: Online action detection with transformers,'' in \emph{Proceedings of the IEEE/CVF International Conference on Computer Vision}, 2021, pp. 7565--7575.

\bibitem{tf}
G.~Bertasius, H.~Wang, and L.~Torresani, ``Is space-time attention all you need for video understanding?'' in \emph{ICML}, vol.~2, no.~3, 2021, p.~4.

\bibitem{lstm}
R.~C. Staudemeyer and E.~R. Morris, ``Understanding lstm--a tutorial into long short-term memory recurrent neural networks,'' \emph{arXiv preprint arXiv:1909.09586}, 2019.

\bibitem{ctrn}
R.~Dai, S.~Das, and F.~Bremond, ``Ctrn: Class-temporal relational network for action detection,'' \emph{arXiv preprint arXiv:2110.13473}, 2021.

\bibitem{thorn}
M.~Guermal, R.~Dai, and F.~Br{\'e}mond, ``Thorn: Temporal human-object relation network for action recognition,'' \emph{arXiv preprint arXiv:2204.09468}, 2022.

\end{thebibliography}


\begin{thebibliography}{1}
\bibitem{CTRN}
Dai, R., Das, S., & Bremond, F. (2021). CTRN: Class-Temporal Relational Network for Action Detection. BMVC 2021
\bibitem{Epic-Kitchen}
 Damen, D., Doughty, H., Farinella, G. M., Fidler, S., Furnari, A., Kazakos, E., ... & Wray, M.,
Scaling egocentric vision: The epic-kitchens dataset
In Proceedings of the European Conference on Computer Vision (ECCV), 2018
\bibitem{STIN}
Materzynska, J., Xiao, T., Herzig, R., Xu, H., Wang, X., & Darrell, T. 
Something-else: Compositional action recognition with spatial-temporal interaction networks In Proceedings of the IEEE/CVF Conference on Computer Vision and Pattern Recognition (pp. 1049-1059).,2020
\bibitem{X3D}
FEICHTENHOFER, Christoph,
X3d: Expanding architectures for efficient video recognition,
Proceedings of the IEEE/CVF Conference on Computer Vision and Pattern Recognition (CVPR), 2020

\bibitem{human-computer}
Xinghao Jiang, Ke Xu, and Tanfeng Sun.,Action recognition scheme based on skeleton representation with ds-lstm network.,IEEE Transactions on Circuits and Systems for Video Technology, 30(7):2129–2140, 2019


\bibitem{video-surveillance}
Tam V Nguyen and Bilal Mirza.,Dual-layer kernel extreme learning machine for action  recognition.Neurocomputing, 2560:123–130,2017


\bibitem{two-stream-1}
Karen Simonyan and Andrew Zisserman.,Two-stream convolutional networks for action recognition in videos.,In Advances in Neural Information Processing Systems, 2014


\bibitem{two-stream-2}
 Christoph Feichtenhofer, Axel Pinz, and Richard P Wildes., Spatiotemporal multiplier networks for video action recognition.,In Proceedings of the IEEE conference on computer vision and pattern recognition, pages 4768–4777,, 2017

\bibitem{two-stream-3}
Christoph Feichtenhofer, Axel Pinz, and Andrew Zisserman.,Convolutional two-stream network fusion for video action recognition.,In Proceedings of the IEEE conference on computer vision and pattern recognition, pages 1933–1941,2016


\bibitem{3D-cnn-1}
Shuiwang Ji, Wei Xu, Ming Yang, and Kai Yu.,3d convolutional neural networks for human action recognition.,IEEE transactions on pattern analysis and machine intelligence, 35(1):221–231,2013


\bibitem{3D-cnn-2}
Joao Carreira and Andrew Zisserman.,Quo vadis, action recognition? a new model and the kinetics dataset.,In proceedings of the IEEE Conference on Computer Vision and Pattern Recognition, pages 6299–6308,2017


\bibitem{3D-cnn-3}
Xiaolong Wang, Ross Girshick, Abhinav Gupta, and Kaiming He.,Non-local neural networks.,In Pro-ceedings of the IEEE conference on computer vision and pattern recognition, pages 7794–7803,,2018


\bibitem{vd1}
Will Kay, Joao Carreira, Karen Simonyan, Brian Zhang, Chloe Hillier, Sudheendra Vijayanarasimhan, Fabio Viola,Tim Green, Trevor Back, Paul Natsev, et al.,The kinetics human action video dataset.,arXiv preprint arXiv:1705.06950,2017



\bibitem{vd2}
Hildegard Kuehne, Hueihan Jhuang, Estíbaliz Garrote, Tomaso Poggio, and Thomas Serre.,Hmdb: a large video database for human motion recognition.,In Proceedings of the IEEE International Conference on Computer Vision, pages 2556–2563. IEEE,2011







\bibitem{vd3}
Andrej Karpathy, George Toderici, Sanketh Shetty, Thomas Leung, Rahul Sukthankar, and Li Fei-Fei.,Large-scale video classification with convolutional neural networks.,In Proceedings of the IEEE conference on Computer Vision and Pattern Recognition, pages 1725–1732,2014




\bibitem{vd4}
Khurram Soomro, Amir Roshan Zamir, and Mubarak Shah.,Ucf101: A dataset of 101 human actions classes from videos in the wild,In Proceedings of the IEEE conference on computer vision and pattern recognition,2012



\bibitem{video-as-graph}
Wang, X., & Gupta, A.,Videos as space-time region graphs,In Proceedings of the European conference on computer vision (ECCV) (pp. 399-417).,2018


\bibitem{TPN}
Yang, C., Xu, Y., Shi, J., Dai, B., & Zhou, B. ,Temporal pyramid network for action recognition,In Proceedings of the IEEE/CVF Conference on Computer Vision and Pattern Recognition (pp. 591-600).,2020


\bibitem{2-1-D}
Du Tran, Heng Wang, Lorenzo Torresani, Jamie Ray, Yann LeCun, and Manohar Paluri.,A closer look at spatiotemporal convolutions for action recognition.,In Proceedings of the IEEE conference on Computer,364 Vision and Pattern Recognition, pages 6450–6459,2018


\bibitem{21D}
Ji Lin, Chuang Gan, and Song Han.,Tsm: Temporal shift module for efficient video understanding.,In Proceedings of the IEEE International Conference on Computer Vision, pages 7083–7093,2019


\bibitem{21-D}
Yan Li, Bin Ji, Xintian Shi, Jianguo Zhang, Bin Kang, and Limin Wang.,Tea: Temporal excitation and aggregation for action recognition.,In Proceedings of the IEEE Conference on Computer Vision and Pattern Recognition, pages 909–918,2020


\bibitem{GCN}
Kipf, T.N., Welling, M,: Semi-supervised classification with graph convolutional
networks.,In: International Conference on Learning Representations (ICLR),2017


\bibitem{Gskel}
Yan, S., Xiong, Y., Lin, D.,Spatial temporal graph convolutional networks for
skeleton-based action recognition., In: AAAI,2018


\bibitem{STGCN}
Ghosh, P., Yao, Y., Davis, L., & Divakaran, A,Stacked spatio-temporal graph convolutional networks for action segmentation.,In Proceedings of the IEEE/CVF Winter Conference on Applications of Computer Vision (pp. 576-585),2020

\bibitem{agcn}
Shi, L., Zhang, Y., Cheng, J., & Lu, H. (2019). Two-stream adaptive graph convolutional networks for skeleton-based action recognition. In Proceedings of the IEEE/CVF conference on computer vision and pattern recognition (pp. 12026-12035).


\bibitem{fastcnn}
Chen, Y., Li, W., Sakaridis, C., Dai, D., & Van Gool, L.,Domain adaptive faster r-cnn for object detection in the wild.,In Proceedings of the IEEE conference on computer vision and pattern recognition (pp. 3339-3348).,2018

\bibitem{UNIK}
Yang, D., Wang, Y., Dantcheva, A., Garattoni, L., Francesca, G., & Bremond, F. (2021). UNIK: A Unified Framework for Real-world Skeleton-based Action Recognition. BMVC2021

\bibitem{few-shot}
Guo, M., Chou, E., Huang, D. A., Song, S., Yeung, S., & Fei-Fei, L.,Neural graph matching networks for fewshot 3d action recognition.,In Proceedings of the European conference on computer vision (ECCV) (pp. 653-669).,2018

\bibitem{LFB}
Wu, C. Y., Feichtenhofer, C., Fan, H., He, K., Krahenbuhl, P., & Girshick, R. (2019). Long-term feature banks for detailed video understanding. In Proceedings of the IEEE/CVF Conference on Computer Vision and Pattern Recognition (pp. 284-293).
\bibitem{RU-LSTM}
Osman, N., Camporese, G., Coscia, P., & Ballan, L. (2021). SlowFast Rolling-Unrolling LSTMs for Action Anticipation in Egocentric Videos. In Proceedings of the IEEE/CVF International Conference on Computer Vision (pp. 3437-3445).
\bibitem{TBN}
Kazakos, E., Nagrani, A., Zisserman, A., & Damen, D. (2019). Epic-fusion: Audio-visual temporal binding for egocentric action recognition. In Proceedings of the IEEE/CVF International Conference on Computer Vision (pp. 5492-5501).
\bibitem{Anet}
Wang, L., & Koniusz, P. (2021, October). Self-supervising action recognition by statistical moment and subspace descriptors. In Proceedings of the 29th ACM International Conference on Multimedia (pp. 4324-4333).
\bibitem{LSC}
\end{thebibliography}
\newpage

%

\end{document}